\documentclass{article}

\usepackage{microtype}
\usepackage{graphicx}
\usepackage{subcaption}
\usepackage{booktabs} % for professional tables

\usepackage{multirow}
\usepackage{makecell}
\usepackage{tabularx}
\usepackage[most]{tcolorbox}

\usepackage{hyperref}

\usepackage[accepted]{icml2026}

\usepackage{amsmath}
\usepackage{amssymb}
\usepackage{mathtools}
\usepackage{amsthm}

\usepackage{algpseudocode}

\usepackage{xcolor}
\usepackage{url}

\usepackage[capitalize,noabbrev]{cleveref}

\theoremstyle{plain}

\theoremstyle{definition}

\theoremstyle{remark}

\usepackage[textsize=tiny]{todonotes}

\icmltitlerunning{DPR-GM for Anomaly Detection in Cyber-Physical Systems}

\begin{document}

\twocolumn[
  \icmltitle{Domain-Prior-Regularized Graph Modeling \\
    for Anomaly Detection in Cyber-Physical Systems}

  \icmlsetsymbol{equal}{*}

  \begin{icmlauthorlist}
    \icmlauthor{Youngseok Hwang}{snu}
    \icmlauthor{Joonsung Kwon}{snu}
    \icmlauthor{Geonwoo Lee}{snu}
    \icmlauthor{Hyunwoo Park}{snu}
  \end{icmlauthorlist}

  \icmlaffiliation{snu}{Graduate School of Data Science, Seoul National University, Seoul, South Korea}

  \icmlcorrespondingauthor{Hyunwoo Park}{hyunwoopark@snu.ac.kr}

  \icmlkeywords{Anomaly Detection, Graph Neural Networks, Time-Series, Sensor Data, Domain Prior}

  \vskip 0.3in
]



\printAffiliationsAndNotice{}  % no special notice (required even if empty)

\begin{abstract}
% Anomaly detection in multivariate sensor time series is essential for industrial monitoring, but existing graph-based methods often struggle in small-scale physical systems where labeled anomalies are scarce and normal
% operation data are limited. In such settings, end-to-end graph learning can capture spurious correlations,
% and purely data-driven adjacency estimation may produce unstable sensor topologies. We propose \textbf{DPR-GM} (Domain-Prior-Regularized Graph Modeling), a forecasting-based anomaly detection framework that incorporates system design knowledge into graph construction.
% DPR-GM uses a binary domain adjacency matrix as a structural gate and modulates the permitted edges using Pearson correlations estimated only from normal
% training data. The anomaly score further incorporates sensor-level reliability weights derived from the coefficient of variation, giving larger influence to sensors that remain stable during nominal operation.
% The graph structure and reliability weights are fixed before model training and do not introduce additional learnable parameters, which reduces the risk of
% label leakage and overfitting to dataset-specific correlations. Experiments on the SKAB benchmark under a strict point-wise evaluation protocol show that DPR-GM improves over graph-based, statistical, and deep learning
% baselines across F1, AUROC, and AUPRC. These results suggest that domain-structured graph priors provide a practical and robust alternative to fully learned graph topology in data-scarce industrial anomaly detection.
Anomaly detection on multivariate sensor time series is critical for industrial monitoring of cyber-physical systems (CPS), where even subtle deviations from normal behavior can indicate process disruption. Recent graph-based approaches have made significant progress, but they often struggle in small-scale physical systems with scarce labeled anomalies and limited normal data. In such settings, graph-based models tend to capture spurious correlations and produce unstable sensor topologies. We propose DPR-GM (\textbf{D}omain-\textbf{P}rior-\textbf{R}egularized \textbf{G}raph \textbf{M}odeling), a forecasting-based framework that incorporates system design knowledge into graph construction. DPR-GM leverages a large language model (LLM) to extract directed physical couplings between sensor pairs from system documentation, which are encoded as a binary domain adjacency matrix serving as a structural gate over sensor relations. This gate is then modulated by Pearson correlations estimated from normal training data. The anomaly score is further weighted by sensor-level reliability derived from the coefficient of variation. All graph and weighting components are fixed prior to training and add no learnable parameters. On the SKAB benchmark, DPR-GM outperforms graph-based, statistical, and deep learning baselines across F1, AUROC, and AUPRC, showing that domain-structured graph priors are a practical alternative to fully learned topologies in data-scarce CPS.
\end{abstract}

% sec/1_Introduction.tex

\section{Introduction}
\label{sec:intro}

Anomaly detection in multivariate sensor time series is a critical task in industrial cyber-physical systems (CPS), such as water treatment plants, water distribution networks, and chemical process control systems~\cite{mathur2016swat, ahmed2017wadi, downs1993plant, skab}. In such systems, early fault or attack identification can prevent costly downtime, safety incidents, and operational disruptions. Because labeled fault examples are scarce in practice, the problem is commonly approached in an unsupervised setting, where models learn normal dynamics from clean operational data and flag deviations at inference time~\cite{ruff2018dsvdd, su2019omni, zhang2019mscred}.

Recent work has shown that modeling \emph{inter-sensor relationships} via Graph Neural Networks (GNNs) substantially improves detection quality by capturing propagation patterns across physically coupled sensors~\cite{deng2021gdn, zhao2020mtadgat, dai2022ganf, liu2025gcad}. However, existing approaches construct the sensor graph either by end-to-end learning~\cite{deng2021gdn, zhang2022grelen} or by thresholding empirical correlations, both of which are unreliable when normal-operation data is limited.
Learned graphs absorb spurious correlations, and correlation-only graphs are destabilized by transient co-movements unrelated to physical causality.

\begin{figure}[h]
  \centering
  \includegraphics[width=1.0\columnwidth]{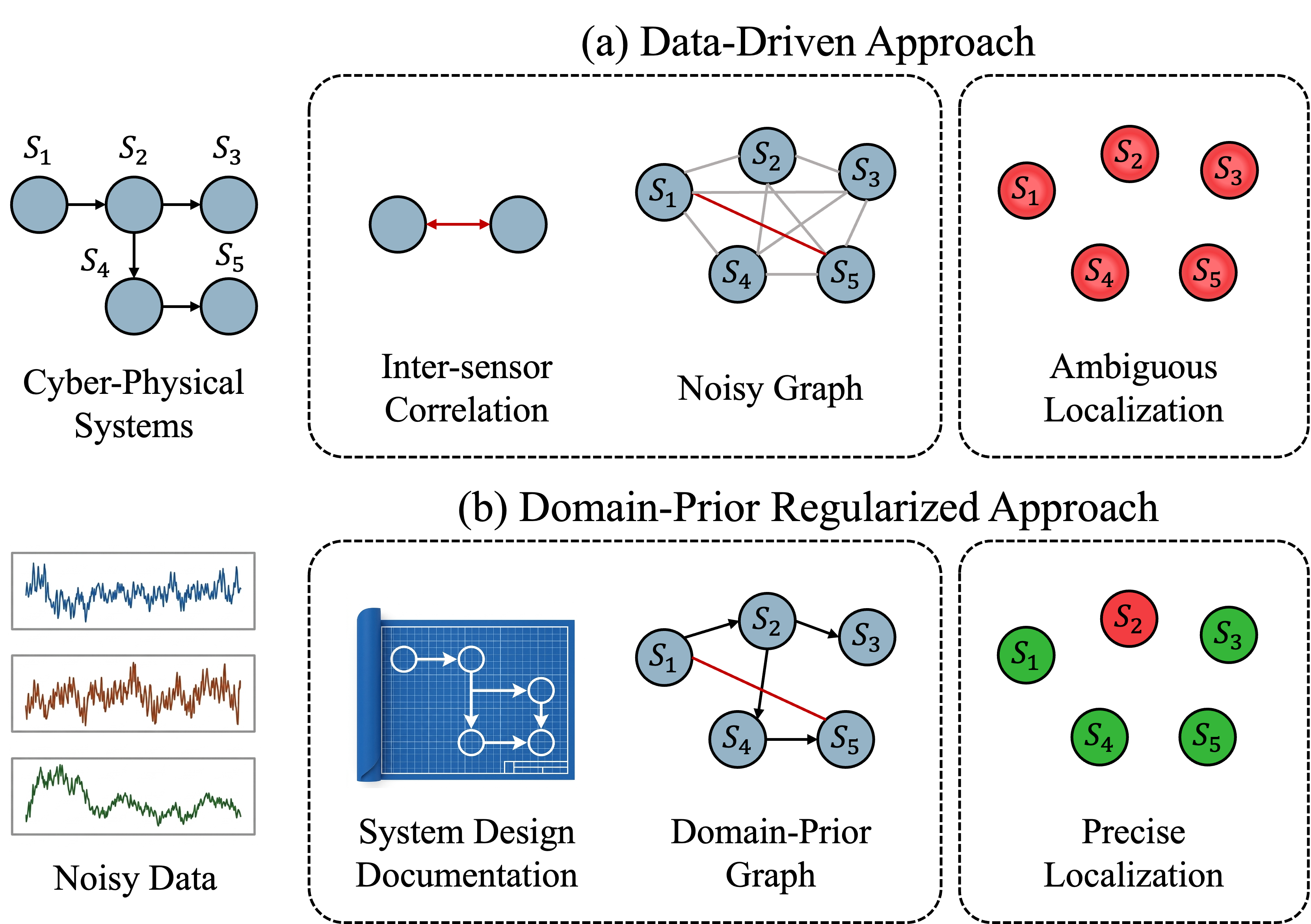}
  \caption{Motivation for domain-prior regularized graph modeling.}
  \label{fig:motivation}
\end{figure}

Figure~\ref{fig:motivation} illustrates how system design documentation can constrain noisy data-driven graphs and support more precise anomaly localization.
In many CPS, the system design already encodes physical sensor coupling, subsystem membership, and output variable roles. This \emph{domain prior} should constrain graph topology before any data-driven modulation is applied, yet current methods rely solely on inter-sensor correlation, producing noisy graphs and ambiguous localization.

We propose \textbf{DPR-GM}
(\textbf{D}omain-\textbf{P}rior-\textbf{R}egularized \textbf{G}raph
\textbf{M}odeling), a framework that uses a binary domain adjacency matrix as a structural gate, suppressing any edge not sanctioned by domain knowledge, while modulating permitted edge weights via Pearson correlation estimated from normal training data. Anomaly scoring further incorporates per-sensor reliability weights derived from the coefficient of variation (CV), giving higher influence to sensors that exhibit stable behavior under normal operation. Neither the graph structure nor the node weights involve any learnable parameters, making the prior entirely grounded in domain knowledge. In this work, we use a large language model (LLM) to automatically extract directed physical couplings from system documentation~\cite{claude_sonnet46}, broadening applicability to systems without manual expert annotation.

We evaluate DPR-GM on the SKAB benchmark~\cite{skab} under the strict point-wise evaluation protocol, the most demanding regime for cyber-physical system anomaly detection. Through systematic ablation over graph construction variants, we demonstrate that combining domain structure with data-driven edge modulation consistently outperforms each information source in isolation, and improves over graph-based, statistical, and deep learning baselines on threshold-free ranking metrics.

We summarize our contributions as follows:
\begin{itemize}
  \item We propose \textbf{DPR-GM}, which gates sensor graph topology using domain knowledge and modulates edge weights via Pearson correlation, with CV-based reliability weighting for anomaly scoring.
  \item We demonstrate on SKAB that DPR-GM outperforms graph-based, statistical, and deep learning baselines on threshold-free ranking metrics under the strict point-wise evaluation protocol.
  \item We show that domain priors in DPR-GM can be automatically extracted from system documentation via LLM, broadening applicability to CPS without manual expert annotation.
\end{itemize}
% sec/2_Related_Work.tex

\section{Related Work}
\label{sec:related}

\subsection{Multivariate Time-Series Anomaly Detection}

Classical approaches to multivariate time-series anomaly detection include statistical process monitoring~\citep{box2015time, hotelling1947multivariate}, PCA-based reconstruction~\citep{shyu2003novel}, one-class classifiers~\citep{scholkopf2001estimating, tax2004support}, and density-based methods such as Local Outlier  Factor~\citep{breunig2000lof} and Isolation Forest~\citep{liu2008isolation}.
These methods are computationally efficient and interpretable, but rely on stationarity or linearity assumptions and struggle to capture nonlinear temporal
dynamics and high-order inter-variable dependencies.

Deep learning methods overcome these limitations by modeling normal temporal dynamics using forecasting, reconstruction, or hybrid objectives. Forecasting-based models detect anomalies when future observations deviate from predicted values. LSTM-NDT~\citep{hundman2018detecting} applies stacked LSTMs with non-parametric dynamic thresholding, while THOC~\citep{shen2020timeseries} introduces a dilated RNN with a temporal hierarchical one-class objective. TimesNet~\citep{wu2023timesnet} reshapes 1D time series into 2D tensors via FFT-based period decomposition to jointly capture intra- and inter-period variations. Reconstruction-based models detect anomalies as poorly reconstructed inputs. LSTM-VAE~\citep{park2018multimodal} models probabilistic temporal dynamics but ignores explicit inter-variable dependencies; OmniAnomaly~\citep{su2019robust} addresses this via stochastic recurrent networks with Gaussian state-space models; USAD~\citep{audibert2020usad} further improves stability through an adversarial autoencoder framework.
Hybrid methods combine both objectives: TranAD~\citep{tuli2022tranad} uses attention-based encoders with adversarial training; Anomaly Transformer~\citep{xu2022anomaly} introduces an Association Discrepancy criterion that exploits the weaker series associations of anomalous points.

Despite their effectiveness, these non-graph methods treat inter-sensor dependencies only implicitly through shared latent representations. In CPS, faults propagate across electrically, mechanically, hydraulically, and thermally coupled subsystems. Without explicit relational structure, models cannot distinguish a physically meaningful dependency change from an independent channel-level fluctuation, and with limited normal-operation data they have little incentive to learn dependency structure that matches physical causal pathways.

\subsection{Graph-Based Multivariate Time-Series Anomaly Detection}

Graph-based approaches address this by representing sensors as nodes and inter-sensor dependencies as edges, enabling GNNs to combine spatial and
temporal modeling~\citep{wu2020comprehensive, jin2024survey}. GDN~\citep{deng2021gdn} learns a sparse sensor graph through embedding similarity and scores anomalies as normalized forecast deviations. MTAD-GAT~\citep{zhao2020multivariate} applies graph attention over both feature and temporal dimensions with a joint forecasting and reconstruction
objective. GReLeN~\citep{zhang2022grelen} infers latent graph structure via variational relational inference, using changes in the inferred structure itself as an
anomaly signal. Another direction treats the sensor graph as a directed or causal structure. 
GANF~\citep{dai2022graph} places a Bayesian network over time series and decomposes the joint density into conditional normalizing flows for density-based anomaly scoring. GCAD~\citep{liu2025gcad} estimates dynamic Granger causality and defines anomalies as deviations from learned causal dependency patterns. DVGCRN~\citep{chen2022dvgcrn} further combines graph learning with variational recurrent modeling to capture fine-grained spatio-temporal dependencies.

However, most graph-based methods rely on data-driven graph construction, which is unreliable when normal-operation data are limited. Learned correlations may reflect coincidental statistical patterns rather than
physical causal pathways, and the resulting graph can be unstable across training runs. DPR-GM addresses this by using a directed domain adjacency matrix as an  explicit structural prior, anchoring the graph in known physical coupling rather than treating it as a free parameter to be learned. DPR-GM also differs from prior methods by incorporating sensor reliability into anomaly scoring via the coefficient of variation rather than learning channel importance end-to-end, consistent with the finding in CATCH~\citep{wu2024catch} that channel importance is critical for multivariate anomaly detection.

% sec/4_Methodology.tex

\section{Methodology}
\label{sec:method}

\begin{figure}[t]
  \centering
  \includegraphics[width=1.0\columnwidth]{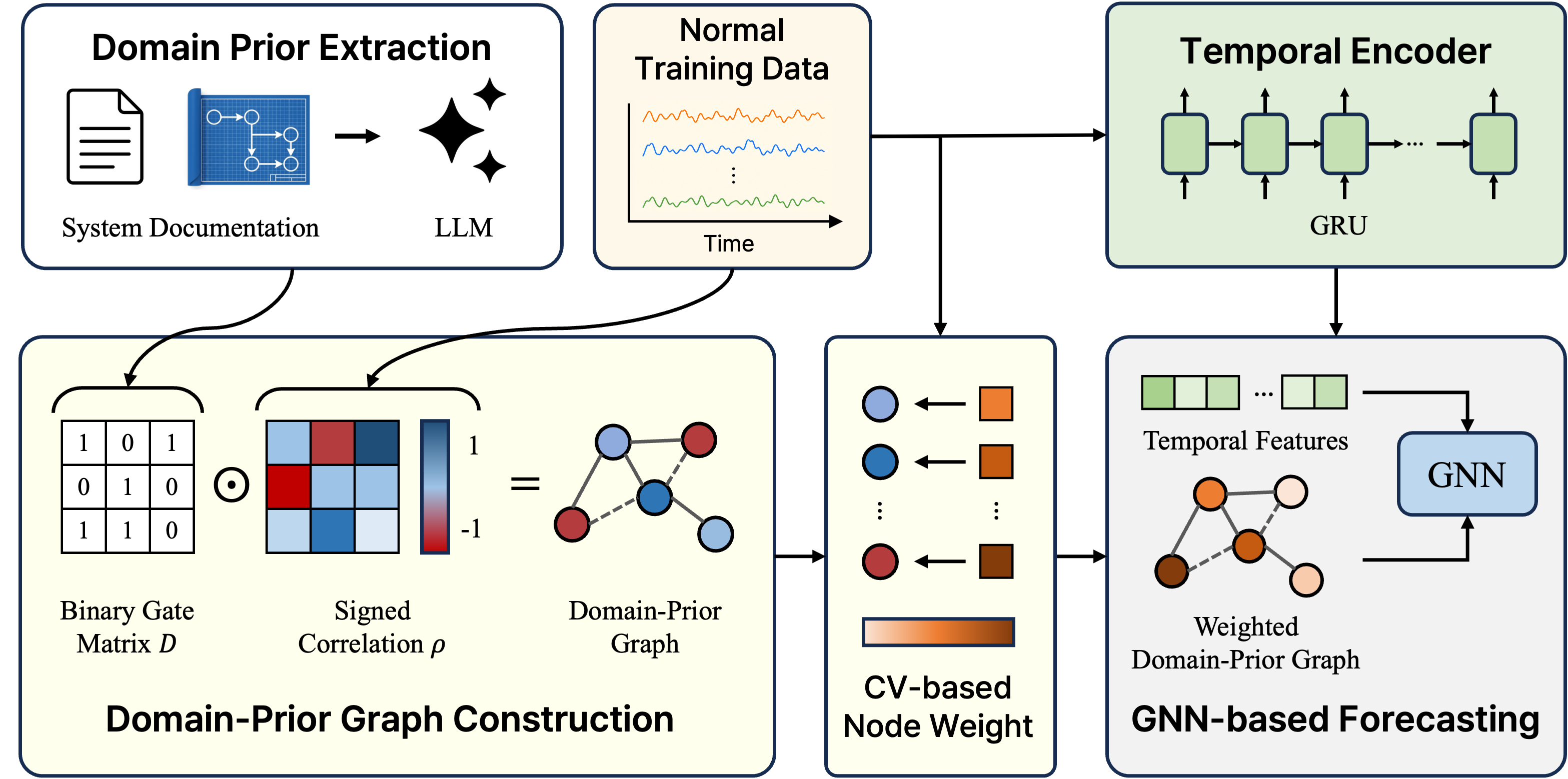}
  \caption{Overview of DPR-GM for domain-prior graph construction and GNN-based forecasting with CV-based node reliability weighting.}
  \label{fig:overview}
\end{figure}

We present \textbf{DPR-GM}, a forecasting-based anomaly detection framework built on the principle that domain knowledge determines which edges may exist, while data statistics determine how strong those edges are.
As illustrated in Figure~\ref{fig:overview}, the framework proceeds in two stages:
(1) \emph{Domain-prior graph construction} uses an LLM to extract directed physical dependencies from system documentation, forming a binary domain adjacency matrix whose edge weights are modulated by signed Pearson correlation estimated from normal training data.
(2) \emph{GNN-based forecasting} encodes sensor time series with a shared temporal encoder and propagates them over the resulting domain-prior graph to generate multi-step predictions. Anomaly scores are further weighted by per-sensor reliability derived from the coefficient of variation.

%%% ---------------------------------------------------------------
\subsection{Domain-Prior-Regularized Graph Construction}
\label{subsec:graph}

\paragraph{Domain prior extraction.}
In a physical sensor system, sensor relationships are governed by the underlying physical processes rather than being arbitrary. An upstream driving variable predictably propagates its effect to a downstream response variable through a known physical pathway, giving rise to a \emph{directed} dependency that is asymmetric in general. Such structural knowledge is encoded in system design documentation before any data is collected, yet current graph-based methods discard
it entirely.

We extract this prior by prompting an LLM with system design documents, sensor metadata, and subsystem descriptions. Specifically, we use Claude Sonnet~\cite{claude_sonnet46} in this work. The full prompt is provided in Appendix~\ref{appendix:prompt}. The prompt instructs the model to identify directed physical couplings between sensor pairs, specifying the source sensor, the target sensor, and the physical mechanism of influence. The output is a structured list of directed sensor pairs $(i \to j)$ that are physically sanctioned, which we use to construct the binary domain adjacency matrix $\mathbf{D}$. Directed structure that cannot be reliably recovered from correlation alone is thus made available to the model before training begins.

\paragraph{Domain adjacency matrix.}
Let $\mathcal{G} = (\mathcal{V}, \mathcal{E})$ denote the sensor graph, where
each node $v_i \in \mathcal{V}$ corresponds to a physical sensor and each
directed edge $(i, j) \in \mathcal{E}$ represents a physically sanctioned
dependency from sensor $i$ to sensor $j$.
We encode this structural knowledge as a binary domain adjacency matrix
$\mathbf{D} \in \{0,1\}^{N \times N}$, where $D_{ij} = 1$ if and only if a
directed physical coupling from sensor $i$ to sensor $j$ is specified by system
design documentation.
All entries outside this specification are set to zero unconditionally,
regardless of any empirical evidence.
The matrix $\mathbf{D}$ is therefore \emph{asymmetric} in general: the presence
of an edge $(i, j)$ does not imply the existence of the reverse edge $(j, i)$.

\paragraph{Correlation modulation.}
Domain knowledge determines the \emph{topology} of the graph, but not the relative importance of individual edges. To modulate edge strength by empirical evidence, we estimate the Pearson correlation matrix $\boldsymbol{\rho} \in [-1,1]^{N \times N}$ from
normal training data $\mathcal{X}_{\text{train}}$ only:
\begin{equation}
  \rho_{ij} = \frac{\mathrm{Cov}(x_i, x_j)}{\sigma_i \sigma_j},
\end{equation}
where $x_i$ denotes the time series of sensor $i$ over the training set. The final DPR-GM adjacency matrix $\mathbf{A}$ is defined as:
\begin{equation}
  A_{ij} = D_{ij} \cdot \bigl(0.5 + 0.5\,\rho_{ij}\bigr).
  \label{eq:adjacency}
\end{equation}

The formulation in Equation~\eqref{eq:adjacency} encodes two properties
simultaneously.
First, $\mathbf{D}$ acts as a \emph{binary gate}: any edge not sanctioned by
domain knowledge receives zero weight regardless of empirical correlation, preventing spurious sensor-pair associations from entering the
graph.
Second, for domain edges, the signed correlation term maps positive, zero, and negative correlations to edge weights above, equal to, and below the neutral baseline of $0.5$, respectively. Table~\ref{tab:edge_cases} summarizes representative boundary cases of the
resulting signed edge-weighting rule.

\begin{table}[h]
  \caption{Edge weight $A_{ij}$ under representative signed-correlation conditions.}
  \label{tab:edge_cases}
  \centering
  \begin{small}
    \begin{tabular*}{\columnwidth}{@{\extracolsep{\fill}}lccc}
      \toprule
      Condition & $D_{ij}$ & $\rho_{ij}$ & $A_{ij}$ \\
      \midrule
      Domain edge, positive correlation & 1 & $1.0$  & $1.0$ \\
      Domain edge, no correlation       & 1 & $0.0$  & $0.5$ \\
      Domain edge, negative correlation & 1 & $-1.0$ & $0.0$ \\
      No domain edge       & 0 & any    & $0.0$ \\
      \bottomrule
    \end{tabular*}
  \end{small}
\end{table}

The use of signed correlation $\rho_{ij}$ encodes an additional semantic in the edge modulation rule. For domain edges, positively correlated sensor pairs receive larger message-passing weights, uncorrelated pairs receive a neutral weight of $0.5$,
and negatively correlated pairs are downweighted, reaching zero when $\rho_{ij}=-1$. This signed modulation makes the edge strength sensitive to the direction of normal-operation co-movement while still respecting the domain gate $\mathbf{D}$.
The $0.5$ baseline ensures that domain edges with no observed correlation remain structurally active, preventing physically motivated edges from being silenced solely by insufficient training statistics. Table~\ref{tab:edge_cases} illustrates representative boundary cases of this signed edge-weighting rule.

% \begin{table}[t]
%   \caption{Edge weight $A_{ij}$ under representative conditions.}
%   \label{tab:edge_cases}
%   \begin{center}
%     \begin{small}
%       \begin{sc}
%         \begin{tabular}{lccc}
%           \toprule
%           Condition & $D_{ij}$ & $|\rho_{ij}|$ & $A_{ij}$ \\
%           \midrule
%           Domain edge, strong correlation & 1 & 1.0 & 1.0 \\
%           Domain edge, weak correlation   & 1 & 0.0 & 0.5 \\
%           No domain edge                  & 0 & any & 0.0 \\
%           \bottomrule
%         \end{tabular}
%       \end{sc}
%     \end{small}
%   \end{center}
% \end{table}

%%% ---------------------------------------------------------------
\subsection{CV-Based Node Reliability Weights}

\begin{figure}[h]
  \centering
  \includegraphics[width=\linewidth]{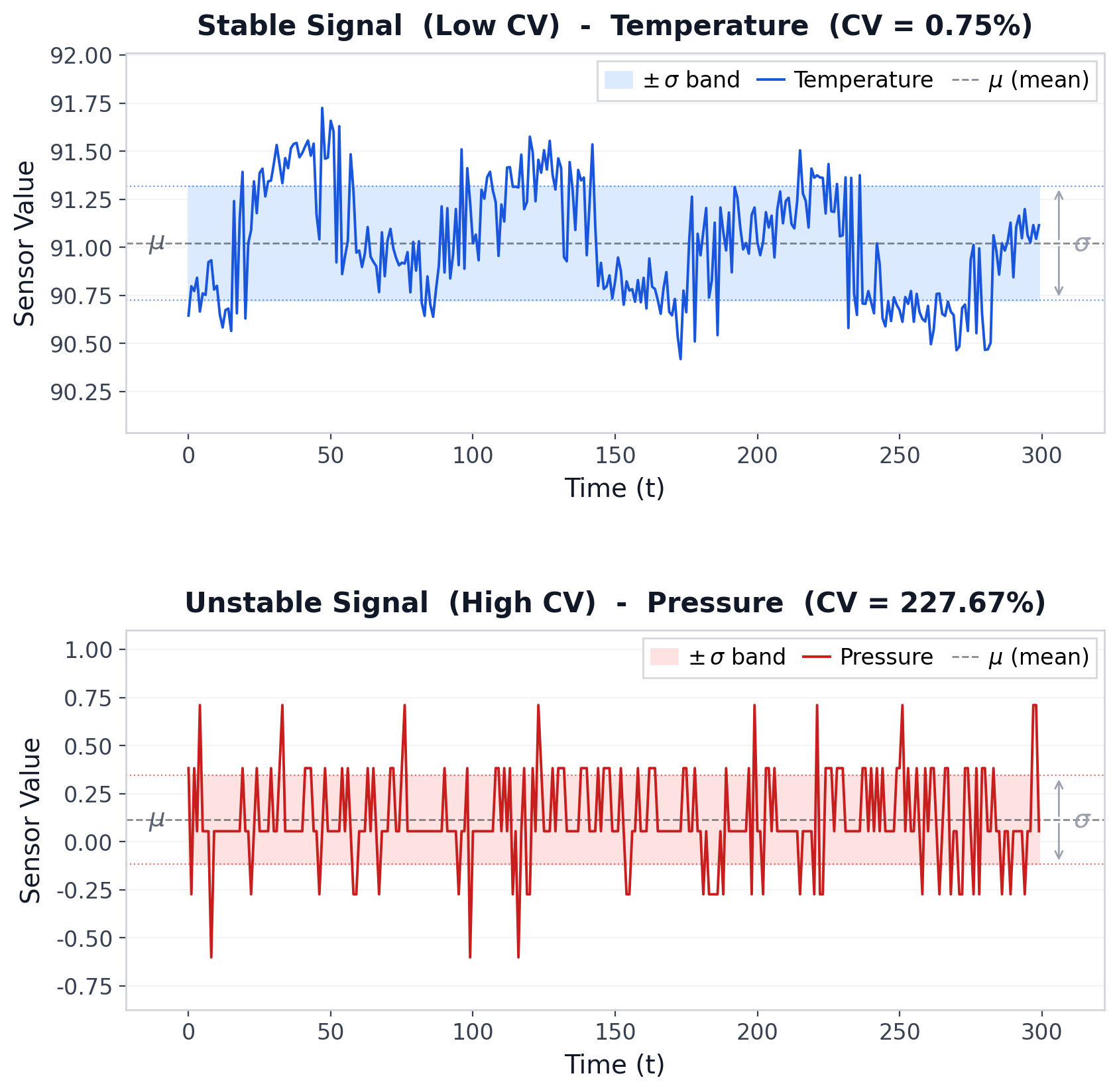}
  \caption{
  Illustration of CV-based node reliability weighting, where stable low-CV sensors receive higher reliability scores than highly variable high-CV sensors.
  }
  \label{fig:cv_node_weight}
\end{figure}

Sensors exhibit heterogeneous variability during normal operation. A sensor with low temporal variability in the normal regime constitutes a reliable reference point: deviations in its readings are more likely to reflect genuine fault conditions than measurement noise.
We operationalize this intuition via the CV:
\begin{equation}
  \mathrm{CV}_i = \frac{\sigma_i}{|\mu_i| + \varepsilon} \times 100,
\end{equation}
and define the reliability score $\boldsymbol{r}$ and normalized node weight $\mathbf{w}$ as:
\begin{equation}
  r_i = \frac{1}{\mathrm{CV}_i + \varepsilon}, \qquad
  w_i = \frac{r_i}{\sum_{j=1}^{N} r_j},
  \label{eq:nodeweight}
\end{equation}
where $\varepsilon > 0$ is a small constant for numerical stability. Figure~\ref{fig:cv_node_weight} visually contrasts low-CV stable signals with high-CV variable signals underlying the proposed reliability weighting.  Both $\mathbf{A}$ and $\mathbf{w}$ are computed exclusively from $\mathcal{X}_{\text{train}}$ and CPS documentation, and remain fixed throughout training and inference without introducing any data-driven parameters into the graph structure.

\label{subsec:nodeweight}

%%% ---------------------------------------------------------------
\subsection{Model Architecture}
\label{subsec:arch}

\paragraph{Input.}
The model receives a sliding window
$\mathbf{X} \in \mathbb{R}^{B \times T \times N}$,
where $B$ is the mini-batch size, $T$ is the window length, and $N$ is the number of sensors.

\paragraph{Temporal encoding.}
A weight-shared GRU is applied independently to each normalized sensor channel.
The input is reshaped from $\mathbb{R}^{B \times T \times N}$ to $\mathbb{R}^{BN \times T \times 1}$ so that all $N$ channels share the same recurrent weights. This parameter-efficient design treats the GRU as a common temporal motif encoder for standardized univariate sensor sequences, which helps reduce overfitting in small-scale datasets.

Let $\mathbf{H} \in \mathbb{R}^{BN \times T \times d_h}$ denote the hidden
sequence produced by the shared GRU. To capture temporal information beyond the final hidden state, we aggregate the hidden sequence using the final state, temporal mean pooling, and temporal max pooling, followed by a linear projection:
\begin{equation}
  \mathbf{h} = \mathrm{Linear}\bigl(
    \mathbf{h}_{\mathrm{last}} \,\|\,
    \mathbf{h}_{\mathrm{mean}} \,\|\,
    \mathbf{h}_{\mathrm{max}}
  \bigr) \in \mathbb{R}^{BN \times d}.
\end{equation}

\paragraph{Sensor identity embedding.}
Because the GRU is shared across channels, sensors with similar normalized temporal patterns may produce similar hidden representations. We therefore concatenate a learnable sensor embedding $\mathbf{e}_i \in \mathbb{R}^{d_e}$ to each temporal representation before graph message passing, allowing the model to
distinguish which physical sensor generated the pattern. The sensor embeddings are repeated across the batch to form $\mathbf{E} \in \mathbb{R}^{BN \times d_e}$:
\begin{equation}
  \mathbf{h} \leftarrow \bigl[\mathbf{h} \,\|\, \mathbf{E}\bigr]
  \in \mathbb{R}^{BN \times (d + d_e)}.
\end{equation}

\paragraph{Graph message passing.}
The DPR-GM adjacency $\mathbf{A}$ is converted to a sparse edge list. For each mini-batch, the same sensor graph is replicated $B$ times and batched by node
offset for a single GNN call. A general message-passing layer first aggregates incoming messages:
\begin{equation}
  \mathbf{M}_{b,i}^{(\ell)}
  =
  \underset{j \in \mathcal{N}_{\mathrm{in}}(i)}{\mathrm{AGGREGATE}}\;
  \mathbf{m}\!\left(
    \mathbf{h}_{b,i}^{(\ell)},
    \mathbf{h}_{b,j}^{(\ell)},
    A_{ji}
  \right),
  \label{eq:message_aggregation}
\end{equation}
and then updates the node representation as
\begin{equation}
  \mathbf{h}_{b,i}^{(\ell+1)}
  =
  \mathrm{UPDATE}\!\left(
    \mathbf{h}_{b,i}^{(\ell)},
    \mathbf{M}_{b,i}^{(\ell)}
  \right).
  \label{eq:node_update}
\end{equation}
Here, $\mathcal{N}_{\mathrm{in}}(i)=\{j \mid A_{ji}>0\}$ denotes the in-neighborhood of node $i$, and $A_{ji}$ denotes the edge weight from source
node $j$ to target node $i$. The function $\mathbf{m}(\cdot)$ denotes the backbone-specific message sent from source node $j$ to target node $i$. It may
apply a linear transformation, edge-weight modulation, or attention-based reweighting before the resulting messages are combined by the permutation-invariant aggregation operator. The specific instantiation of
Equations~\eqref{eq:message_aggregation}--\eqref{eq:node_update} is determined
by the choice of backbone.

We consider GCN~\cite{kipf2016gcn}, GraphSAGE~\cite{hamilton2017sage},
GAT~\cite{velivckovic2017gat}, and Graph Transformer~\cite{shi2020gt}
backbones. In our implementation, GCN and weighted GraphSAGE explicitly consume the
DPR-GM edge weights during message passing. GAT and Graph Transformer use only the nonzero graph support induced by $\mathbf{A}$, while their aggregation weights are learned from node features. Unless otherwise stated, we use weighted GraphSAGE as the default backbone.

\paragraph{Forecasting head.}
The node representations are projected to $k$-step-ahead predictions:
\begin{equation}
    \hat{\mathbf{Y}} = \mathrm{reshape}\bigl(
      \mathrm{Linear}(\mathbf{h})
    \bigr) \in \mathbb{R}^{B \times k \times N}.
\end{equation}
The model is trained to minimize the mean squared error (MSE) between
$\hat{\mathbf{Y}}$ and the corresponding future ground-truth window
$\mathbf{Y} \in \mathbb{R}^{B \times k \times N}$.

%%% ---------------------------------------------------------------
\subsection{Anomaly Scoring}
\label{subsec:scoring}

At inference, the per-sensor forecast error for each window is computed as:
\begin{equation}
  e_{b,i} = \frac{1}{k}\sum_{\tau=1}^{k}
    \bigl(\hat{Y}_{b,\tau,i} - Y_{b,\tau,i}\bigr)^2.
\end{equation}

To obtain a distribution-free normalized error, IQR normalization is applied using statistics fitted on the training set:
\begin{equation}
  \tilde{e}_{b,i} = \max\!\left(0,\;
    \frac{e_{b,i} - \mathrm{median}(e_i^{\mathrm{train}})}
         {\mathrm{IQR}(e_i^{\mathrm{train}})}\right).
\end{equation}
% The clamp at zero removes sub-baseline errors that would otherwise suppress the
% final score.

The window-level anomaly score aggregates the weighted per-sensor errors using a combination of a mean term and a max term: 
\begin{equation}
  s_b = (1 - \alpha)\frac{1}{N}\sum_{i=1}^{N} w_i\,\tilde{e}_{b,i}
    \;+\; \alpha \max_{i}\,w_i\,\tilde{e}_{b,i},
  \label{eq:score}
\end{equation}
where $\alpha = 0.2$ and $w_i$ are the CV-based reliability weights from
Equation~\eqref{eq:nodeweight}.
The mean term captures distributed multi-sensor anomalies, while the max term retains sensitivity to localized high-amplitude deviations in a single sensor.

\begin{table*}[!t]
  \caption{Graph-based method comparison on SKAB (strict point-wise).
  Mean (std) over five repeated runs.
  \textbf{Bold} / \underline{underline}: best / second-best in \texttt{overall}. See Section~\ref{subsec:baselines} for method references.
  }
  \label{tab:graph_comparison}
  \begin{center}
    \begin{scriptsize}
      \begin{sc}
        \begin{tabularx}{0.9\textwidth}{ll *{6}{>{\centering\arraybackslash}X}}
          \toprule
          Method & Fault & Precision & Recall & Prac-F1 & AUROC & AUPRC & BF-F1 \\
          \midrule
          \multirow{4}{*}{GDN}
            & valve1  & 0.4190{\tiny(0.3873)} & 0.0003{\tiny(0.0004)} & 0.0007{\tiny(0.0007)} & 0.4947{\tiny(0.0098)} & 0.3436{\tiny(0.0063)} & 0.5166{\tiny(0.0008)} \\
            & valve2  & 0.0000{\tiny(0.0000)} & 0.0000{\tiny(0.0000)} & 0.0000{\tiny(0.0000)} & 0.5270{\tiny(0.0112)} & 0.3604{\tiny(0.0060)} & 0.5252{\tiny(0.0057)} \\
            & other   & 0.3555{\tiny(0.3185)} & 0.1143{\tiny(0.1262)} & 0.1493{\tiny(0.1515)} & 0.5788{\tiny(0.0872)} & 0.4672{\tiny(0.0775)} & 0.5388{\tiny(0.0221)} \\
            \cmidrule(lr){2-8}
            & \bfseries overall & \underline{0.4750}{\tiny(0.2713)} & 0.0460{\tiny(0.0507)} & 0.0753{\tiny(0.0795)} & 0.5314{\tiny(0.0314)} & 0.3820{\tiny(0.0228)} & 0.5223{\tiny(0.0047)} \\
          \midrule
          \multirow{4}{*}{GReLeN}
            & valve1  & 0.3147{\tiny(0.1668)} & 0.0996{\tiny(0.1256)} & 0.1180{\tiny(0.1433)} & 0.4836{\tiny(0.0278)} & 0.3492{\tiny(0.0149)} & 0.5261{\tiny(0.0001)} \\
            & valve2  & 0.2199{\tiny(0.1799)} & 0.1102{\tiny(0.1346)} & 0.1283{\tiny(0.1533)} & 0.4990{\tiny(0.0171)} & 0.3601{\tiny(0.0099)} & 0.5318{\tiny(0.0005)} \\
            & other   & 0.5154{\tiny(0.1758)} & 0.0566{\tiny(0.0523)} & 0.0947{\tiny(0.0830)} & 0.5116{\tiny(0.0633)} & 0.3899{\tiny(0.0698)} & 0.5308{\tiny(0.0001)} \\
            \cmidrule(lr){2-8}
            & \bfseries overall & \textbf{0.4866}{\tiny(0.1872)} & 0.0836{\tiny(0.0877)} & 0.1161{\tiny(0.1138)} & 0.4971{\tiny(0.0337)} & 0.3666{\tiny(0.0348)} & 0.5286{\tiny(0.0001)} \\
          \midrule
          \multirow{4}{*}{MTAD-GAT}
            & valve1  & 0.3493{\tiny(0.0000)} & 1.0000{\tiny(0.0000)} & 0.5178{\tiny(0.0000)} & 0.6050{\tiny(0.0058)} & 0.5234{\tiny(0.0128)} & 0.5246{\tiny(0.0001)} \\
            & valve2  & 0.3518{\tiny(0.0000)} & 1.0000{\tiny(0.0000)} & 0.5205{\tiny(0.0000)} & 0.5898{\tiny(0.0171)} & 0.4423{\tiny(0.0142)} & 0.5277{\tiny(0.0061)} \\
            & other   & 0.4402{\tiny(0.0011)} & 0.9174{\tiny(0.0064)} & 0.5949{\tiny(0.0008)} & 0.7506{\tiny(0.0090)} & 0.6852{\tiny(0.0103)} & 0.6051{\tiny(0.0065)} \\
            \cmidrule(lr){2-8}
            & \bfseries overall & 0.3795{\tiny(0.0002)} & \textbf{0.9669}{\tiny(0.0026)} & \underline{0.5450}{\tiny(0.0004)} & \underline{0.5794}{\tiny(0.0027)} & \underline{0.4566}{\tiny(0.0075)} & \underline{0.5476}{\tiny(0.0018)} \\
          \midrule
          \multirow{4}{*}{\textbf{DPR-GM (Ours)}}
            & valve1  & 0.3480{\tiny(0.0000)} & 1.0000{\tiny(0.0000)} & 0.5163{\tiny(0.0000)} & 0.8315{\tiny(0.0154)} & 0.8225{\tiny(0.0162)} & 0.7194{\tiny(0.0165)} \\
            & valve2  & 0.3518{\tiny(0.0000)} & 1.0000{\tiny(0.0000)} & 0.5205{\tiny(0.0000)} & 0.8616{\tiny(0.0187)} & 0.8590{\tiny(0.0131)} & 0.7817{\tiny(0.0097)} \\
            & other   & 0.4995{\tiny(0.0428)} & 0.8890{\tiny(0.0306)} & 0.6377{\tiny(0.0267)} & 0.8131{\tiny(0.0008)} & 0.7638{\tiny(0.0007)} & 0.6781{\tiny(0.0057)} \\
            \cmidrule(lr){2-8}
            & \bfseries overall & 0.3922{\tiny(0.0086)} & \underline{0.9555}{\tiny(0.0123)} & \textbf{0.5560}{\tiny(0.0066)} & \textbf{0.7256}{\tiny(0.0058)} & \textbf{0.7165}{\tiny(0.0084)} & \textbf{0.6063}{\tiny(0.0094)} \\
          \bottomrule
        \end{tabularx}
      \end{sc}
    \end{scriptsize}
  \end{center}
  \vskip -0.1in
\end{table*}

% sec/5_Experiment.tex

\section{Experiments}
\label{sec:experiment}

%%% ------------------------------------------------------------------
\subsection{Dataset}
\label{subsec:dataset}

We evaluate DPR-GM on the \textbf{Skoltech Anomaly Benchmark (SKAB)}~\cite{skab}, a multivariate time series benchmark collected  from a physical water pump testbed under controlled industrial fault conditions. The testbed consists of a water pump, an electric motor, an inverter, two electromagnetic valves, and a set of sensors measuring current, voltage,
vibration (two axes), pressure, flow rate, temperature, and thermocouple readings ($N = 8$ sensors in total).
The dataset contains 34 fault scenarios across three splits: \texttt{valve1} (18,130 time steps), \texttt{valve2} (4,312 time steps), and \texttt{other} (14,920 time steps), totaling 37,362 time steps in the \texttt{overall} split. The anomaly fraction is approximately 35\% across all splits.
We use the \texttt{overall} split as the primary evaluation target.

\subsection{Evaluation Protocol and Experimental Setup}
\label{subsec:protocol}

We adopt a strict point-wise evaluation protocol, where each time step is scored independently and classified as anomalous only if its anomaly score exceeds a threshold. We do not use point adjustment, event-level matching, or window-based correction, since such post-processing can obscure whether a method detects anomalies at the correct time step. This conservative protocol
is motivated by recent critiques showing that common benchmark and evaluation choices in time-series anomaly detection can lead to unreliable comparisons and an inflated impression of progress \citep{wu2021current}. We therefore treat strict point-wise performance as the primary evaluation criterion.

Let TP, FP, TN, and FN denote the point-wise entries of the confusion matrix.
We report Precision, Recall, F1 score, Matthews Correlation Coefficient
(\textbf{MCC}), area under the ROC curve (\textbf{AUROC}), and area under the
precision-recall curve (\textbf{AUPRC}). MCC is defined as:
\begin{equation}
  \mathrm{MCC}
  =
  \frac{\mathrm{TP}\cdot\mathrm{TN} - \mathrm{FP}\cdot\mathrm{FN}}
  {\sqrt{\scriptstyle(\mathrm{TP}+\mathrm{FP})(\mathrm{TP}+\mathrm{FN})
         (\mathrm{TN}+\mathrm{FP})(\mathrm{TN}+\mathrm{FN})}}.
\end{equation}
AUROC and AUPRC integrate threshold-free ranking quality over all decision boundaries where $\mathrm{TPR}=\mathrm{TP}/(\mathrm{TP}+\mathrm{FN})$,
$\mathrm{FPR}=\mathrm{FP}/(\mathrm{FP}+\mathrm{TN})$, and
$\mathrm{Prec}$, $\mathrm{Rec}$ are precision and recall parameterized by threshold. For thresholded metrics, the anomaly threshold is selected by Peak-over-Threshold (PoT) using only training scores, and we report the resulting practical F1 (Prac-F1), reflecting deployment where test labels are unavailable. As a diagnostic upper bound, we also report the oracle best-F1 (BF-F1), obtained by sweeping all thresholds on the test set.

\paragraph{Experimental setup.}
Unless otherwise specified, we use an input window length of $T=30$, a forecasting horizon of $k=10$, $N=8$ sensors, and a batch size of $B=256$. All models are trained on normal training windows and evaluated on held-out fault scenarios using the same preprocessing, windowing, thresholding, and metric computation pipeline. We report mean and standard deviation over five repeated runs. Details of the computational environment are provided in  Appendix~\ref{appendix:environment}.

%%% ------------------------------------------------------------------
\subsection{Baselines}
\label{subsec:baselines}
We compare DPR-GM with two groups of baselines. Graph-based baselines test the advantage of the proposed domain-prior graph over learned or inferred sensor
graphs. Non-graph baselines assess the benefit of explicit physical dependency modeling over standard statistical and deep anomaly detection models without
graph structure.

\paragraph{Graph-based baselines.}
We re-implement GDN~\cite{deng2021gdn}, which learns a sparse sensor graph from embedding similarity; MTAD-GAT~\cite{zhao2020mtadgat}, which applies feature-
and time-oriented graph attention with forecasting and reconstruction losses; and GReLeN~\cite{zhang2022grelen}, which infers latent relations via variational graph learning.

\paragraph{Non-graph baselines.}
We also compare against statistical methods (T$^2$+Q, T-squared, MSET), density-based detection (ISF~\cite{liu2008isolation}), and deep
reconstruction-based models (Conv-AE, Vanilla AE, Vanilla LSTM~\cite{hochreiter1997lstm},
LSTM-AE, LSTM-VAE~\cite{park2018multimodal}, MSCRED~\cite{zhang2019mscred}).

%% ---- TABLE 3: Broader comparison (two column) ----
\begin{table*}[t]
  \caption{Comparison with statistical and deep learning baselines on SKAB
  \texttt{overall} (strict point-wise, PoT threshold). Mean (std).
  \textbf{Bold}: best overall. \underline{Underline}: best among non-graph baselines.}
  \label{tab:broader_comparison}
  \centering
  \begin{scriptsize}
    \begin{sc}
      \setlength{\tabcolsep}{4pt}
      \renewcommand{\arraystretch}{1.1}
      \begin{tabular*}{\textwidth}{@{\extracolsep{\fill}}lccccccc}
        \toprule
        Method & Precision & Recall & Prac-F1 & MCC & AUROC & AUPRC & BF-F1 \\
        \midrule
        T$^2$+Q
          & 0.3737 & 0.9693 & 0.5394
          & 0.1571 & 0.6656 & 0.6133 & 0.5490 \\
        T-squared
          & 0.3745 & 0.9705 & 0.5404
          & 0.1614 & 0.6737 & 0.6246 & 0.5580 \\
        MSET
          & 0.3695{\scriptsize(0.0017)} & 0.8422{\scriptsize(0.0138)} & 0.5136{\scriptsize(0.0042)}
          & 0.0823{\scriptsize(0.0091)} & 0.6388{\scriptsize(0.0003)} & 0.5908{\scriptsize(0.0006)} & 0.5336{\scriptsize(0.0004)} \\
        ISF
          & 0.3452{\scriptsize(0.0034)} & 0.5272{\scriptsize(0.1538)} & 0.4077{\scriptsize(0.0591)}
          & $-$0.0086{\scriptsize(0.0056)} & 0.5105{\scriptsize(0.0013)} & 0.3494{\scriptsize(0.0020)} & 0.5316{\scriptsize(0.0007)} \\
        Conv-AE
          & 0.3567{\scriptsize(0.0025)} & 0.9942{\scriptsize(0.0067)} & 0.5250{\scriptsize(0.0022)}
          & 0.0911{\scriptsize(0.0155)} & 0.6296{\scriptsize(0.0239)} & 0.5428{\scriptsize(0.0476)} & 0.5491{\scriptsize(0.0026)} \\
        Vanilla AE
          & \underline{0.3751}{\scriptsize(0.0013)} & 0.9697{\scriptsize(0.0003)} & \underline{0.5409}{\scriptsize(0.0013)}
          & \underline{0.1631}{\scriptsize(0.0054)} & 0.6374{\scriptsize(0.0015)} & 0.5634{\scriptsize(0.0018)} & 0.5481{\scriptsize(0.0022)} \\
        Vanilla LSTM
          & 0.3627{\scriptsize(0.0010)} & 0.9821{\scriptsize(0.0023)} & 0.5298{\scriptsize(0.0010)}
          & 0.1149{\scriptsize(0.0052)} & 0.6359{\scriptsize(0.0013)} & 0.5623{\scriptsize(0.0036)} & 0.5558{\scriptsize(0.0003)} \\
        LSTM-AE
          & 0.3557{\scriptsize(0.0009)} & \underline{0.9986}{\scriptsize(0.0011)} & 0.5246{\scriptsize(0.0009)}
          & 0.0919{\scriptsize(0.0062)} & 0.6337{\scriptsize(0.0012)} & 0.5608{\scriptsize(0.0019)} & 0.5525{\scriptsize(0.0009)} \\
        LSTM-VAE
          & 0.3550{\scriptsize(0.0010)} & \textbf{0.9993}{\scriptsize(0.0015)} & 0.5238{\scriptsize(0.0011)}
          & 0.0873{\scriptsize(0.0093)} & 0.6694{\scriptsize(0.0019)} & 0.6194{\scriptsize(0.0041)} & 0.5507{\scriptsize(0.0013)} \\
        MSCRED
          & 0.3690{\scriptsize(0.0002)} & 0.9721{\scriptsize(0.0004)} & 0.5349{\scriptsize(0.0002)}
          & 0.1381{\scriptsize(0.0008)} & \underline{0.7029}{\scriptsize(0.0186)} & \underline{0.6809}{\scriptsize(0.0312)} & \underline{0.5809}{\scriptsize(0.0309)} \\
        \midrule
        \textbf{DPR-GM (Ours)}
          & \textbf{0.3922}{\scriptsize(0.0086)} & 0.9555{\scriptsize(0.0123)} & \textbf{0.5560}{\scriptsize(0.0066)}
          & \textbf{0.2129}{\scriptsize(0.0181)} & \textbf{0.7256}{\scriptsize(0.0058)} & \textbf{0.7165}{\scriptsize(0.0084)} & \textbf{0.6063}{\scriptsize(0.0094)} \\
        \bottomrule
      \end{tabular*}
    \end{sc}
  \end{scriptsize}
  \vskip -0.1in
\end{table*}
%%% ------------------------------------------------------------------
\subsection{Results}
\label{subsec:results}

% Table~\ref{tab:graph_comparison} compares DPR-GM with graph-based anomaly detection methods across all SKAB fault splits. Table~\ref{tab:broader_comparison} compares DPR-GM with statistical and deep learning baselines on the \texttt{overall} split. 
% Table~\ref{tab:backbone_ablation} evaluates four GNN architectures within the same DPR-GM framework. Table~\ref{tab:ablation} reports an ablation study over five repeated runs that isolates the contribution of each design component.  

Table~\ref{tab:graph_comparison} compares DPR-GM with graph-based anomaly detection methods across all SKAB fault splits. Table~\ref{tab:broader_comparison} compares DPR-GM with statistical and deep learning baselines on the \texttt{overall} split. 
Table~\ref{tab:backbone_ablation} evaluates four GNN architectures within the same DPR-GM framework, where the SAGE row corresponds to the DPR-GM used in the main comparison. Table~\ref{tab:ablation} reports an ablation study over five repeated runs that isolates the contribution of each design component.

%%% ------------------------------------------------------------------
% \subsection{Analysis}
% \label{subsec:analysis}

\paragraph{Graph-based comparison (Table~\ref{tab:graph_comparison}).}
% GDN and GReLeN collapse on SKAB, achieving near-zero F1 scores. Their high run-to-run variance indicates that end-to-end graph learning is unstable when normal-operation data are limited. MTAD-GAT recovers competitive F1 through near-universal recall, but its substantially weaker AUROC (0.579) and AUPRC
% (0.457) reveal that its anomaly scores carry limited discriminative structure, suggesting that its graph-attention mechanism is tuned toward high recall rather than toward producing well-separated anomaly scores. DPR-GM closes this gap, with the largest gains appearing on the threshold-free metrics (AUPRC $+0.260$ over MTAD-GAT), indicating that DPR-GM's score distribution itself is sharper, not merely that its operating point is better tuned. Overall, DPR-GM attains the best F1, AUROC, AUPRC, and BF-F1 simultaneously, evidence that domain-structured priors yield discriminative anomaly scores rather than recall-biased ones.
GDN and GReLeN collapse on SKAB, achieving near-zero F1 scores of 0.075 and 0.116, respectively, with large run-to-run variance. This failure mode is
characteristic of end-to-end graph learning under limited normal-operation data. Without sufficient observations to stabilize the learned topology, the model captures spurious correlations, yielding graph structures that vary across runs and degrade detection quality. Because neither method is anchored by domain knowledge, any transient co-movement in the training window can enter the graph as a false edge, leading to inconsistent message-passing behavior at inference.

MTAD-GAT avoids this collapse by deriving aggregation weights from node features via attention rather than learning the graph topology end-to-end, recovering competitive Prac-F1 (0.545) through near-universal recall. However, its AUROC (0.579) and AUPRC (0.457) remain substantially weaker, revealing that high Prac-F1 here is a threshold artifact rather than a reflection of genuine score discrimination. A method that flags nearly every time step as anomalous will inevitably achieve high recall, but its anomaly scores carry little information about the true
degree of anomalousness at each point.

DPR-GM addresses both failure modes by grounding the graph in domain knowledge before any data-driven modulation is applied. The binary domain adjacency matrix eliminates spurious edges unconditionally, providing the stable topology that learned methods cannot recover from limited data. Signed Pearson correlation then modulates permitted edge weights by empirical co-movement, so the graph encodes both structural and statistical information without conflating the two. The gains are consistent across all three fault splits. On \texttt{valve1} and
\texttt{valve2}, DPR-GM achieves AUROC above 0.83 and 0.86 respectively, substantially outperforming MTAD-GAT, indicating that domain-prior structure is particularly effective for valve-type faults where physical coupling between sensors is well-defined. On \texttt{other}, DPR-GM still achieves AUROC 0.813 and AUPRC 0.764, suggesting that the domain prior generalizes beyond the fault types present during graph construction. The consistency of these improvements further supports the view that DPR-GM's gains stem from a structurally sounder graph rather than from overfitting to a particular anomaly pattern. Overall, DPR-GM attains the best Prac-F1, AUROC, AUPRC, and BF-F1 simultaneously, evidence that domain-structured priors yield discriminative anomaly scores rather than recall-biased ones.

% \begin{table}[h]
%   \caption{Ablation study on SKAB \texttt{overall} over 20 runs (strict point-wise).
%   Mean (std). \textbf{Bold} / \underline{underline}: best / second-best.
%   }
%   \label{tab:ablation}
%   \centering
%   \begin{scriptsize}
%     \begin{sc}
%       \setlength{\tabcolsep}{4pt}
%       \begin{tabular*}{\columnwidth}{@{\extracolsep{\fill}}lccc}
%         \toprule
%         Variant & F1 & AUROC & AUPRC \\
%         \midrule
%         No-graph
%           & 0.5506{\scriptsize(0.0022)}
%           & 0.7141{\scriptsize(0.0100)}
%           & 0.6970{\scriptsize(0.0178)} \\
%         Corr-only
%           & \underline{0.5497}{\scriptsize(0.0022)}
%           & \underline{0.7154}{\scriptsize(0.0111)}
%           & \underline{0.7001}{\scriptsize(0.0194)} \\
%         Domain-only
%           & 0.5482{\scriptsize(0.0037)}
%           & 0.7070{\scriptsize(0.0101)}
%           & 0.6859{\scriptsize(0.0174)} \\
%         DPR-GM (Ours)
%           & \textbf{0.5514}{\scriptsize(0.0048)}
%           & \textbf{0.7259}{\scriptsize(0.0089)}
%           & \textbf{0.7169}{\scriptsize(0.0139)} \\
%         \bottomrule
%       \end{tabular*}
%     \end{sc}
%   \end{scriptsize}
%   \vskip -0.1in
% \end{table}

\paragraph{Non-graph comparison (Table~\ref{tab:broader_comparison}).}
% Among non-graph baselines, Vanilla AE achieves the highest F1 (0.541) and MCC
% (0.163), while MSCRED leads on threshold-free metrics with AUROC 0.703, AUPRC
% 0.681, and BF-F1 0.581. DPR-GM improves over all non-graph baselines, achieving the highest F1 (0.556), MCC (0.213), AUROC (0.726), AUPRC (0.717), and BF-F1 (0.606). Compared with MSCRED, DPR-GM improves AUROC by $+0.023$, AUPRC by $+0.036$, and BF-F1 by $+0.025$. Although classical statistical methods such as T-squared and T$^2$+Q maintain recall near 0.97, their lower AUROC and AUPRC indicate weaker ranking quality. Overall, DPR-GM provides the strongest balance between thresholded detection performance and threshold-free anomaly score discrimination.
Among non-graph baselines, Vanilla AE attains the strongest 
threshold-dependent performance, while MSCRED leads on all 
threshold-free metrics. DPR-GM improves over every non-graph baseline on every reported metric except recall, with AUROC and AUPRC gains of $+0.023$ and $+0.036$ over MSCRED. The MCC improvement over Vanilla AE ($+0.050$) is particularly informative: because
MCC penalizes class-imbalance-driven strategies more strictly than F1, this gain suggests that DPR-GM's decision boundary is genuinely better calibrated rather than simply exploiting the skewed base rate of anomalies. In contrast, classical statistical methods sustain near-saturated recall yet trail substantially on threshold-free metrics, the signature of over-detection.

\paragraph{Backbone comparison (Table~\ref{tab:backbone_ablation}).}
We evaluate four GNN backbones within the same DPR-GM framework to identify the most effective aggregation strategy for the domain-prior graph. In our implementation, GCN and SAGE explicitly consume DPR-GM edge weights during message passing, whereas GAT and GT use only the nonzero graph support and learn attention weights from node features. SAGE achieves the best performance across every reported metric while maintaining low run-to-run variance. GCN is the close second on the threshold-free metrics (AUROC and AUPRC), but it exhibits the highest variance among all backbones on both ($0.027$ and $0.037$), indicating a less stable exploitation of the prior. GT edges out GCN on Prac-F1 by a narrow margin and is the most stable backbone on that metric, though its threshold-free scores trail SAGE. The gap between GCN and SAGE, both edge-weight-aware, suggests that SAGE's separation of self- and neighbor-transformations interacts more favorably with sparse domain-prior graphs than GCN's symmetric normalization. Among the two support-only backbones, GT remains competitive with the edge-weight-aware backbones while GAT performs substantially worse than all three alternatives, indicating that the benefit of the domain-prior graph depends on the specific aggregation mechanism rather than on edge-weight consumption alone. Overall, these results indicate that explicitly consuming domain-prior edge weights during message passing, as in SAGE, yields the strongest and most stable exploitation of the prior, while the choice of aggregation scheme further modulates how well the prior is exploited.

% GT follows as a close second on point estimates but exhibits notably higher run-to-run variance. GT's attention mechanism, which is computed from node features rather than domain-prior edge weights, appears to be more sensitive to variations in
% input representations across runs, trading stability for representational flexibility. SAGE achieves the best performance across every reported metric. The gap between GCN and SAGE suggests that SAGE's separation of self- and neighbor-transformations interacts more favorably with sparse domain-prior graphs than GCN's symmetric normalization. GAT performs substantially worse than all three alternatives since GAT differs from the edge-weight-aware backbones primarily in discarding the prior weights and re-deriving attention from node features. Overall, these results indicate that explicitly consuming domain-prior edge weights during message passing is more effective than re-learning interaction strength from node features, and that among edge-weight-aware backbones, the choice of aggregation scheme further modulates how well the prior is exploited.

%% ---- TABLE 4: Backbone comparison (one column) ----
\begin{table}[h]
    \caption{Backbone comparison on SKAB \texttt{overall} over five repeated runs under the strict point-wise protocol.
  Mean (std). \textbf{Bold} / \underline{underline}: best / second-best. SAGE: GraphSAGE. GT: Graph Transformer.}
  \label{tab:backbone_ablation}
  \centering
  \begin{scriptsize}
    \setlength{\tabcolsep}{3pt}
    \renewcommand{\arraystretch}{1.1}
    \begin{tabular*}{\columnwidth}{@{\extracolsep{\fill}}lcccc}
      \toprule
      Backbone & Edge-W & Prac-F1 & AUROC & AUPRC \\
      \midrule
        GCN
          & Yes
          & 0.5471{\scriptsize(0.0021)}
          & \underline{0.7145}{\scriptsize(0.0268)}
          & \underline{0.6981}{\scriptsize(0.0371)} \\
        GAT
          & No
          & 0.5236{\scriptsize(0.0029)}
          & 0.6091{\scriptsize(0.0042)}
          & 0.5023{\scriptsize(0.0054)} \\
        SAGE
          & Yes
          & \textbf{0.5565}{\scriptsize(0.0024)}
          & \textbf{0.7261}{\scriptsize(0.0115)}
          & \textbf{0.7188}{\scriptsize(0.0168)} \\
        GT
          & No
          & \underline{0.5508}{\scriptsize(0.0006)}
          & 0.7140{\scriptsize(0.0201)}
          & 0.6959{\scriptsize(0.0347)} \\
      \bottomrule
    \end{tabular*}
  \end{scriptsize}
  \vskip -0.1in
\end{table}

\paragraph{Ablation study (Table~\ref{tab:ablation}).}
We evaluate the contribution of Edge-W, the signed correlation-based edge weighting, and Node-W, the CV-based node reliability weighting, while keeping the backbone fixed to weighted GraphSAGE. Across five repeated runs, the full DPR-GM achieves the best Prac-F1, AUROC, and AUPRC, and each ablated variant trails this configuration on every metric we report. Removing Edge-W produces a consistent drop across all three metrics, indicating that signed correlation modulation contributes meaningfully to ranking quality even when the domain topology is preserved. Removing Node-W yields a similarly consistent degradation across all three
metrics, confirming that CV-based reliability weighting provides information complementary to the graph structure alone. Removing both components produces the weakest ranking performance, with the lowest AUROC and AUPRC in the ablation, showing that the domain-prior topology by itself is not sufficient for the strongest ranking quality. Overall, both Edge-W and Node-W consistently contribute to DPR-GM, and their combination produces the most reliable balance of thresholded 
detection quality and threshold-free ranking performance.

%% ---- TABLE 5: Ablation study (one column) ----
\begin{table}[h]
  \caption{Ablation study on SKAB \texttt{overall} over five repeated runs under the strict point-wise protocol.
  Mean (std). \textbf{Bold} / \underline{underline}: best / second-best.
  Edge-W denotes signed correlation-based edge weighting, and Node-W denotes CV-based node reliability weighting.
  }
  \label{tab:ablation}
  \centering
  \begin{scriptsize}
    \begin{tabular*}{\columnwidth}
    {@{\extracolsep{\fill}}lccc}
      \toprule
      Variant & Prac-F1 & AUROC & AUPRC \\
      \midrule
      w/o Edge-W \& Node-W
        & \underline{0.5547}{\scriptsize(0.0040)}
        & 0.7091{\scriptsize(0.0169)}
        & 0.6899{\scriptsize(0.0292)} \\
      w/o Edge-W
        & 0.5518{\scriptsize(0.0031)}
        & \underline{0.7209}{\scriptsize(0.0110)}
        & \underline{0.7113}{\scriptsize(0.0159)} \\
      w/o Node-W
        & 0.5539{\scriptsize(0.0023)}
        & 0.7175{\scriptsize(0.0152)}
        & 0.7028{\scriptsize(0.0270)} \\
      DPR-GM
        & \textbf{0.5565}{\scriptsize(0.0024)}
        & \textbf{0.7261}{\scriptsize(0.0115)}
        & \textbf{0.7188}{\scriptsize(0.0168)} \\
      \bottomrule
    \end{tabular*}
  \end{scriptsize}
  \vskip -0.1in
\end{table}
\section{Conclusion}
\label{sec:conclusion}

We presented DPR-GM, a forecasting-based anomaly detection framework that constructs sensor graphs from domain knowledge rather than fully data-driven graph learning. DPR-GM uses a binary domain adjacency matrix extracted via an LLM from system design documentation to gate edge existence, Pearson correlation to modulate
permitted edge strengths, and CV-based node reliability weights to reflect sensor stability in anomaly scoring.
All graph and reliability priors are fixed before model training and introduce no additional learnable parameters, making the framework well-suited for
data-scarce CPS where overfitting to limited normal-operation statistics is a practical concern. Experiments on the SKAB benchmark under a strict point-wise protocol show that DPR-GM outperforms graph-based, statistical, and deep learning baselines across F1, AUROC, AUPRC, MCC, and BF-F1, demonstrating that explicit domain priors provide a robust and stable alternative to learned graph topology in data-scarce
CPS.

\section{Limitations and Future Work}
\label{sec:limitation}

DPR-GM is limited by the granularity of available domain knowledge.
In SKAB, the documentation provides subsystem membership and signal-flow
direction, but not precise causal strengths between sensor pairs.
Thus, DPR-GM captures coarse structural constraints rather than
fine-grained physical causality. We also validate DPR-GM on a single CPS
benchmark using a single LLM, Claude Sonnet 4.6~\cite{claude_sonnet46}.
Future work will improve LLM-based prior extraction through cross-LLM
validation, prompt robustness analysis, confidence-aware edge extraction, and evaluation across more diverse CPS datasets.

Beyond CPS anomaly detection, many scientific discovery domains generate
sensor-based spatio-temporal time-series data together with system
documentation, experimental protocols, or domain descriptions. DPR-GM suggests
a broader direction in which LLMs transform such documentation into explicit structural priors for graph-based learning. Such priors may support downstream tasks such as forecasting, anomaly detection, imputation, and sensor-state estimation in domains including climate, neuroscience, transportation, maritime systems, and autonomous laboratories. This points toward documentation-grounded AI systems that incorporate symbolic or causal priors into scientific time-series modeling, rather than relying only on correlations learned from limited observations.

% % In the unusual situation where you want a paper to appear in the
% % references without citing it in the main text, use \nocite

\bibliography{ref}
\bibliographystyle{icml2026}

%%%%%%%%%%%%%%%%%%%%%%%%%%%%%%%%%%%%%%%%%%%%%%%%%%%%%%%%%%%%%%%%%%%%%%%%%%%%%%%
%%%%%%%%%%%%%%%%%%%%%%%%%%%%%%%%%%%%%%%%%%%%%%%%%%%%%%%%%%%%%%%%%%%%%%%%%%%%%%%
% APPENDIX
%%%%%%%%%%%%%%%%%%%%%%%%%%%%%%%%%%%%%%%%%%%%%%%%%%%%%%%%%%%%%%%%%%%%%%%%%%%%%%%
%%%%%%%%%%%%%%%%%%%%%%%%%%%%%%%%%%%%%%%%%%%%%%%%%%%%%%%%%%%%%%%%%%%%%%%%%%%%%%%
\newpage
\appendix

\appendix
\onecolumn

\section{LLM Prompt for Domain Prior Extraction}
\label{appendix:prompt}

The following prompt is used to extract directed sensor dependencies from system design documentation using Claude Sonnet 4.6~\cite{claude_sonnet46}. The output is parsed into a structured list of directed sensor pairs $(i \to j)$, which is used to construct the binary domain adjacency matrix $\mathbf{D}$.

\begin{tcolorbox}[
  enhanced,
  breakable,
  colback=white,
  colframe=black!70,
  colbacktitle=black!75,
  coltitle=white,
  title={\large\textbf{System Prompt for Domain Prior Extraction ($N=8$)}},
  fonttitle=\bfseries,
  boxrule=1.2pt,
  arc=3mm,
  outer arc=3mm,
  left=10pt,
  right=10pt,
  top=10pt,
  bottom=10pt,
  toptitle=6pt,
  bottomtitle=6pt,
  boxed title style={
    sharp corners,
    colback=black!75,
    colframe=black!75,
  }
]
\normalfont\small

You are a physical systems expert specializing in sensor-based monitoring of industrial equipment. Given a natural language description of a physical sensor system, your task is to identify all directed physical dependencies between sensor pairs based on the underlying physical processes and system design.

\medskip

\noindent\textbf{Definition.}
A directed dependency from sensor A to sensor B, written as A $\rightarrow$ B, means that a change in the physical quantity measured by A causally and predictably influences the physical quantity measured by B through a known physical mechanism, such as electrical $\rightarrow$ mechanical, mechanical $\rightarrow$ hydraulic, hydraulic $\rightarrow$ thermal, or mechanical $\rightarrow$ mechanical propagation. The dependency is asymmetric: A $\rightarrow$ B does not imply B $\rightarrow$ A.

\medskip

\noindent\textbf{System description:}

\medskip

\noindent\texttt{\{SYSTEM\_DESCRIPTION\}}

\medskip

\noindent\textbf{At each extraction step, you must:}
\begin{itemize}
  \item Identify all sensors in the description.
  \item For each sensor, record its name, the physical quantity it measures, and the subsystem it belongs to: electrical, mechanical, hydraulic, or thermal.
  \item Determine whether each ordered sensor pair has a directed physical dependency based on the system design.
  \item Include only dependencies that are physically motivated by the system design.
  \item Do not include sensor pairs that merely co-move statistically without a physical causal pathway.
\end{itemize}

\medskip

\noindent\textbf{Physical pathways to consider:}
\begin{itemize}
  \item \textit{Electrical $\rightarrow$ Mechanical}: voltage or current driving motor torque, rotation, vibration, or speed.
  \item \textit{Mechanical $\rightarrow$ Hydraulic}: shaft rotation, pump impeller motion, or mechanical actuation driving flow rate or pressure.
  \item \textit{Hydraulic $\rightarrow$ Thermal}: fluid flow, pressure, or circulation affecting temperature or heat dissipation.
  \item \textit{Mechanical $\rightarrow$ Mechanical}: vibration, imbalance, or shaft displacement propagating through connected components.
\end{itemize}

\medskip

\noindent\textbf{For each directed dependency, provide:}
\begin{itemize}
  \item The source sensor.
  \item The physical quantity measured by the source sensor.
  \item The target sensor.
  \item The physical quantity measured by the target sensor.
  \item A one-sentence explanation of the physical mechanism.
\end{itemize}

\medskip

\noindent\textbf{Output format.}
Return the complete result as a JSON list. Do not include any preamble, explanation, or markdown formatting outside the JSON.

\begin{verbatim}
[
  {
    "source": "<sensor name>",
    "source_quantity": "<physical quantity>",
    "target": "<sensor name>",
    "target_quantity": "<physical quantity>",
    "mechanism": "<one-sentence physical explanation>"
  },
  ...
]
\end{verbatim}

\end{tcolorbox}

\section{Computational Environment}
\label{appendix:environment}

All experiments were conducted on an Ubuntu 24.04.4 LTS server equipped with two AMD EPYC 7502 32-core CPUs, 2.0 TiB RAM, and four NVIDIA GeForce RTX 3090
GPUs with 24 GB memory each. The server used NVIDIA driver 580.126.09 and CUDA 13.0.
%%%%%%%%%%%%%%%%%%%%%%%%%%%%%%%%%%%%%%%%%%%%%%%%%%%%%%%%%%%%%%%%%%%%%%%%%%%%%%%
%%%%%%%%%%%%%%%%%%%%%%%%%%%%%%%%%%%%%%%%%%%%%%%%%%%%%%%%%%%%%%%%%%%%%%%%%%%%%%%

\end{document}